\documentclass{article}
\usepackage{spconf,amsmath,graphicx}
\usepackage{booktabs} 
\usepackage{siunitx} 
\usepackage{hyperref}

\title{Multi-stage Learning for Radar Pulse Activity Segmentation}
\name{Zi Huang$^{\star \dagger}$, Akila Pemasiri$^{\star}$, Simon Denman$^{\star}$, Clinton Fookes$^{\star}$, Terrence Martin$^{\dagger}$}
\address{$^{\star}$Queensland University of Technology, Australia \\ $^{\dagger}$Revolution Aerospace, Australia}
\begin{document}
%
\maketitle
\begin{abstract}
Radio signal recognition is a crucial function in electronic warfare. Precise identification and localisation of radar pulse activities are required by electronic warfare systems to produce effective countermeasures. Despite the importance of these tasks, deep learning-based radar pulse activity recognition methods have remained largely underexplored. While deep learning for radar modulation recognition has been explored previously, classification tasks are generally limited to short and non-interleaved IQ signals, limiting their applicability to military applications. To address this gap, we introduce an end-to-end multi-stage learning approach to detect and localise pulse activities of interleaved radar signals across an extended time horizon. We propose a simple, yet highly effective multi-stage architecture for incrementally predicting fine-grained segmentation masks that localise radar pulse activities across multiple channels. We demonstrate the performance of our approach against several reference models on a novel radar dataset, while also providing a first-of-its-kind benchmark for radar pulse activity segmentation.

\end{abstract}
%


\vspace{-5pt}
\section{Introduction}
\label{sec:intro}
\vspace{-5pt}
Radar activity recognition is a fundamental capability of cognitive electronic warfare (CEW) \cite{haigh_cognitive_2021-1}. It encompasses critical sub-functions, such as the detection and classification of unknown radar pulse activities hidden within a low signal-to-noise ratio (SNR) environment. These sub-functions are essential for generating highly accurate pulse descriptor words (PDWs) from the raw signal. A PDW is a data structure used by the radar systems community which provides a common format for representing the value for key signal attributes, such as pulse width (PW) and pulse repetition interval (PRI). Identifying these values is a critical step in any effort to deploy countermeasures against radar threats \cite{robertson2019practical}. Deriving accurate PDWs, therefore, requires precise identification and localisation of radar pulses, which can be complicated by their existence across a long time horizon and the interleaving of multiple pulses in a contested setting.

Contemporary deep learning models \cite{oshea_over--air_2018,vila2019deep,jagannath_multi-task_2021-1, huang2023multi} applied to radio emitter classification and characterisation have been shown to achieve exceptional performance in recent years, however deep learning-based radar pulse activity recognition is an emerging field and thus remains largely underexplored. While similar tasks from adjacent domains, such as speaker diarisation \cite{zhang2019fully}, biomedical signal processing \cite{dissanayake2023multi}, and image semantic segmentation \cite{long2015fully, newell2016stacked} provide a foundational basis to develop robust and high resolution segmentation models, there exist a domain gap in which there is a shortage of publicly available radar datasets with appropriate characteristics to support the development of deep learning models for radar pulse activity segmentation.

Radio datasets, such as RadioML \cite{oshea_over--air_2018},  RadarComms \cite{jagannath_multi-task_2021-1}, and RadChar \cite{huang2023multi} exist in the public domain, however they are not suited for the task of semantic segmentation of radar pulse activities for two key reasons. First, existing datasets do not provide sample-wise annotations. This information is crucial for determining temporal occupancy (e.g., PW) within a given signal. Secondly, existing datasets are limited to non-interleaved and short-duration IQ signals, while realistic radar pulse activities can co-exist and generally occur over an extended time horizon. This second issue is particularly challenging, and put simply, requires fine-grained multi-channel semantic segmentation which is not possible using traditional approaches based on energy detection \cite{kirubahini2020optimal} and pulse correlation \cite{cheng2021enhanced,ge2019improved}. Separately, over-segmentation errors \cite{ishikawa2021alleviating, farha2019ms} can arise due to an imbalance of class activities. Therefore, careful refinement of channel-wise predictions is necessary to predict continuous and smooth activity intervals, a characteristic of real-world radar pulses.

To address these gaps, this paper introduces a multi-stage learning approach which accurately segments pulse activities for interleaved radar signals across an extended time horizon. Our main contributions are threefold. First, we release an open-source dataset\footnote{The download link to our radar pulse activity segmentation dataset can be accessed at: \url{https://github.com/abcxyzi/RadSeg}} containing radar signals with complex interleaving characteristics and long IQ sequences. Secondly, we introduce a simple, yet highly effective end-to-end multi-stage architecture to perform sample-wise signal classification on raw IQ data without requiring expert feature engineering \cite{vila2019deep,logue2019expert}. Finally, we establish a first-of-its-kind benchmark for radar pulse activity segmentation and demonstrate the competitive performance of our multi-stage architecture.


\section{Proposed Method}
\label{sec:method}
\vspace{-5pt}

\subsection{RadSeg Dataset}
\label{ssec:dataset}
\vspace{-5pt}
We introduce a new radar pulse activity dataset (RadSeg) for semantic segmentation. RadSeg builds upon \cite{huang2023multi} and contains $5$ radar signal classes. These include coherent unmodulated pulses (CPT), Barker codes, polyphase Barker codes, Frank codes, and linear frequency-modulated (LFM) pulses. Code lengths of up to $13$ and $16$ are considered for Barker and Frank codes, respectively. Unlike other datasets \cite{oshea_over--air_2018,jagannath_multi-task_2021-1,huang2023multi}, RadSeg contains long-duration signals, each with $32,768$ complex baseband IQ samples ($\Vec{x}_{\text{i}} + j \Vec{x}_{\text{q}}$), compared to $512$ samples provided by RadChar \cite{huang2023multi}. The sampling rate used in RadSeg is $3.2$ \si{\mega\hertz}, which yields a $10.24$ $\si{\milli\second}$ signal duration and a temporal resolution of $0.3125$ $\si{\micro\second}$ per sample. This resolution is chosen to sufficiently capture realistic PWs and PRIs of typical pulsed radar systems \cite{robertson2019practical}.

To generate unique radar pulse activities, several signal parameters are selected and incrementally sampled from uniform distributions to create random unique signal permutations. Importantly, we allow the radar signals to interleave freely in order to model the temporal characteristics of a typical electronic warfare environment \cite{robertson2019practical}. The signal parameters include PW ($t_{\text{pw}}$), PRI ($t_{\text{pri}}$), time of arrival of the first pulse ($t_{\text{toa}}$), number of pulses ($n_{\text{p}}$), and number of signal classes present ($n_{\text{c}}$). The bounds selected for $t_{\text{pw}}$, $t_{\text{pri}}$, $t_{\text{toa}}$, $n_{\text{p}}$, and $n_{\text{c}}$ are $10-100$ $\si{\micro\second}$, $320-5120$ $\si{\micro\second}$, $0-5120$ $\si{\micro\second}$, $2-16$, and $1-5$, respectively, and we uniformly sample from these ranges to create each radar signal class.

\begin{figure}[tb]
\begin{minipage}[b]{1\linewidth}
  \centering
  \centerline{\includegraphics[width=1\linewidth]{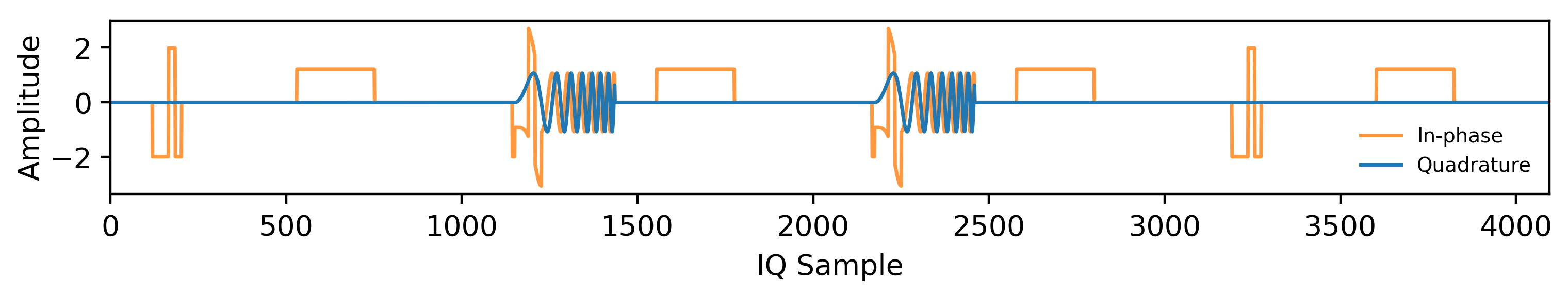}}
  \centerline{(a) Baseband IQ signal interval}\medskip
\end{minipage}
\begin{minipage}[b]{1\linewidth}
  \centering
  \centerline{\includegraphics[width=1\linewidth]{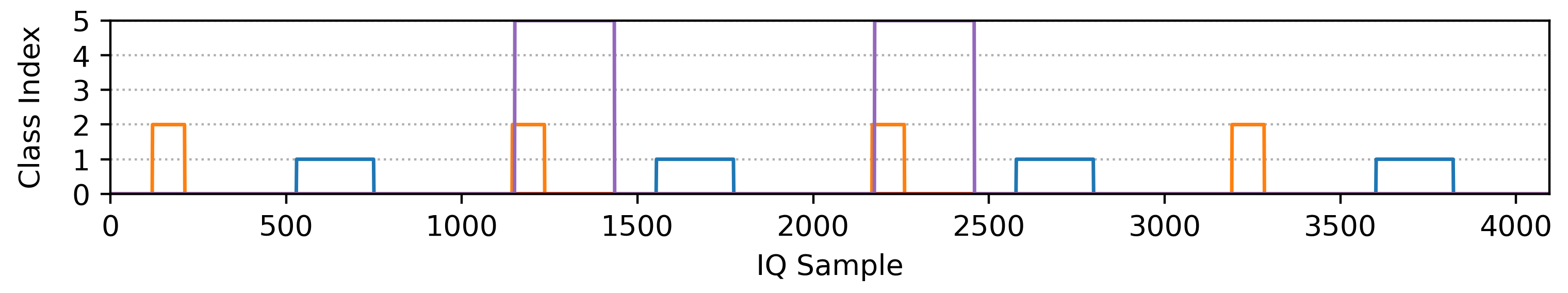}}
  \centerline{(b) Channel-wise segmentation masks}\medskip
\end{minipage}
\caption{RadSeg excerpt with the AWGN channel removed. The class indices $1$, $2$, $3$, $4$, and $5$ correspond to the signal classes CPT, Barker, polyphase Barker, Frank, and LFM, respectively. Here, Barker and LFM pulses are interleaved.}
\label{fig:radar_waveform}
\end{figure}

We generate a total of $80,000$ unique radar signals and provide the dataset in three parts. The training set contains $60,000$ signals, while the validation and test set each contain $10,000$ signals. Additive white Gaussian noise (AWGN) is added each signal to simulate varying SNR settings. We sample SNR from a uniform distribution to produce signals that fall within $-20$ and $20$ \si{\deci\bel} at a resolution of $0.5$ \si{\deci\bel}. Sample-wise ground-truth annotations are provided as $5\times N$ binary segmentation masks where $N$ is the length of the IQ sequence. Each of the $5$ channel masks represents a signal class, where a binary value of $1$ indicates the signal is present at the corresponding sample position. An example from the dataset is shown in Figure \ref{fig:radar_waveform}.

\begin{figure}[htb]
    \centering
    \includegraphics[width=1\linewidth]{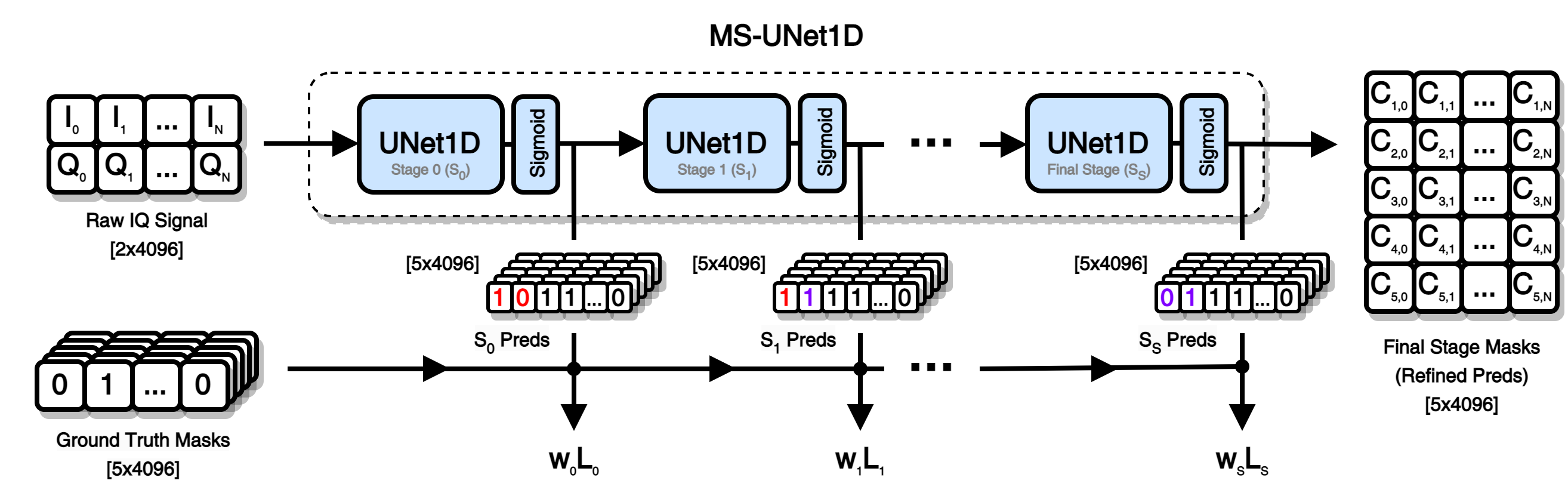}
    \caption{The proposed MS-UNet1D multi-stage learning architecture designed for radar pulse activity segmentation, with channel-wise mask predictions being incrementally refined across its stages $\text{S}_0, \text{S}_1, ..., \text{S}_s$.}
\label{fig:msl_net}
\end{figure}


\vspace{-5pt}
\subsection{Segmentation Models}
\label{ssec:ml_models}
\vspace{-5pt}
We develop temporal semantic segmentation models to establish a benchmark for radar pulse activity segmentation. Our baseline model is a modified UNet \cite{ronneberger2015u} adapted for 1D operations. UNet1D consists of repeated applications of $1\times 3$ convolutions, each followed by a ReLU and $1\times 2$ max pooling, with stride $1$ at each step in the contracting path. Each step in the expansive path consists of repeated upsampling of the feature map using a $1\times 2$ up-convolution and concatenation with the corresponding feature map from the contracting path. Unlike the original architecture, we use padded convolutions to preserve the spatial information of the features at each step, and ensure segmentation masks of the same length as the input signal are produced. The final layer consists of a $1\times 1$ convolution that maps the $64\times N$ feature vectors to $5\times N$ segmentation masks as the final output. The number of output channels can be increased to accommodate additional signal classes. We also apply batch normalisation prior to each ReLU in both the contracting and expansive paths to improve training stability.

To benchmark against the baseline model, we implement MS-TCN \cite{farha2019ms} and MS-TCN++ \cite{li2020ms}, which are both competitive architectures for fine-grained semantic segmentation tasks \cite{lea2017temporal}. We follow the original implementations to adapt these models for our task. For MS-TCN, we use $10$ dilated $1\times 3$ convolutions at each stage. For MS-TCN++, we use $11$ dual dilated $1\times 3$ convolutions in the prediction generation stage, $3$ refinement stages each with $10$ dilated $1\times 3$ convolutions. For both models, the final layer consists of a $1\times 1$ convolution that maps $512\times N$ feature vectors to $5\times N$ segmentation masks as the final output. Unlike the original implementations, we do not apply a softmax activation along the feature dimension of the last layer in order to preserve independent channel activations. This is because multiple signal classes can independently co-exist.

\vspace{-5pt}
\subsection{Multi-stage Learning for Pulse Segmentation}
\label{ssec:multi_stage}
\vspace{-5pt}
To accurately detect and localise pulse activities, a segmentation model is required to consistently extract fine-grained continuous signal features from noise. While task-optimised architectures like the UNet of \cite{ronneberger2015u} utilise high resolution features to produce precise predictions, over-segmentation errors \cite{ishikawa2021alleviating, farha2019ms} can occur if there exists an imbalance of activities in the training data which may cause the model to fluctuate between predictions, or exhibit bias towards certain activities. This is a challenge in electronic warfare where the occurrence of specific activities may be rare. To address this issue, we introduce a multi-stage learning approach to incrementally refine channel-wise mask predictions by sequentially stacking multiple segmentation models. Conceptually, this approach is akin to learning the channel-wise matched filters of the signal at each stage and refining them in subsequent stages.

Multi-stage learning has been shown to be effective at reducing over-segmentation errors in similar tasks \cite{newell2016stacked,farha2019ms,li2020ms}. Motivated by the success of this approach, we introduce a simple, yet effective multi-stage UNet1D (MS-UNet1D) model for precise radar pulse activity segmentation. The proposed model, shown in Figure \ref{fig:msl_net}, consists of a sequential stack of identical UNet1D stages. The first stage ($\text{S}_0$) takes a raw $2\times N$ signal and predicts an initial $5\times N$ mask. The subsequent stage then takes this mask and refines it for the next stage. The loss is computed at the output of each stage during training to minimise the sample-wise dissimilarity between the predicted mask and the actual mask. We introduce a multi-stage loss ($\mathcal{L}_{\text{msl}}$) function given by (\ref{eq:ms_loss}) to evaluate the performance of the multi-stage model during training. Joint optimisation of the multi-stage model is achieved by minimising the total multi-stage loss function as follows
\vspace{-5pt}

\begin{equation}
\label{eq:ms_loss}
    \mathcal{L}_{\text{msl}}(\theta_{\text{0}}, ..., \theta_{s}) = \sum_{i=0}^{s} w_{i}\mathcal{L}_{i}(\theta_{i}),
\end{equation}

\vspace{-3.5pt}
\begin{equation}
\label{eq:opt}
    \operatorname*{argmin}_{\theta_{\text{0}}, ..., \theta_{s}} \mathcal{L}_{\text{msl}}(\theta_{\text{0}}, ...,  \theta_{s}),
\end{equation}

\noindent where each stage is parameterised by stage-specific model parameters ($\theta_{\text{0}}, ...,  \theta_{s}$) and is optimised using binary cross-entropy loss (BCE). The coefficients ($w_i$) of stage-specific losses are hyperparameters. To reduce the number of experimental permutations, we set $w_i$ to $1$.

\begin{figure}[htb]
    \centering
    \includegraphics[width=1.0\linewidth]{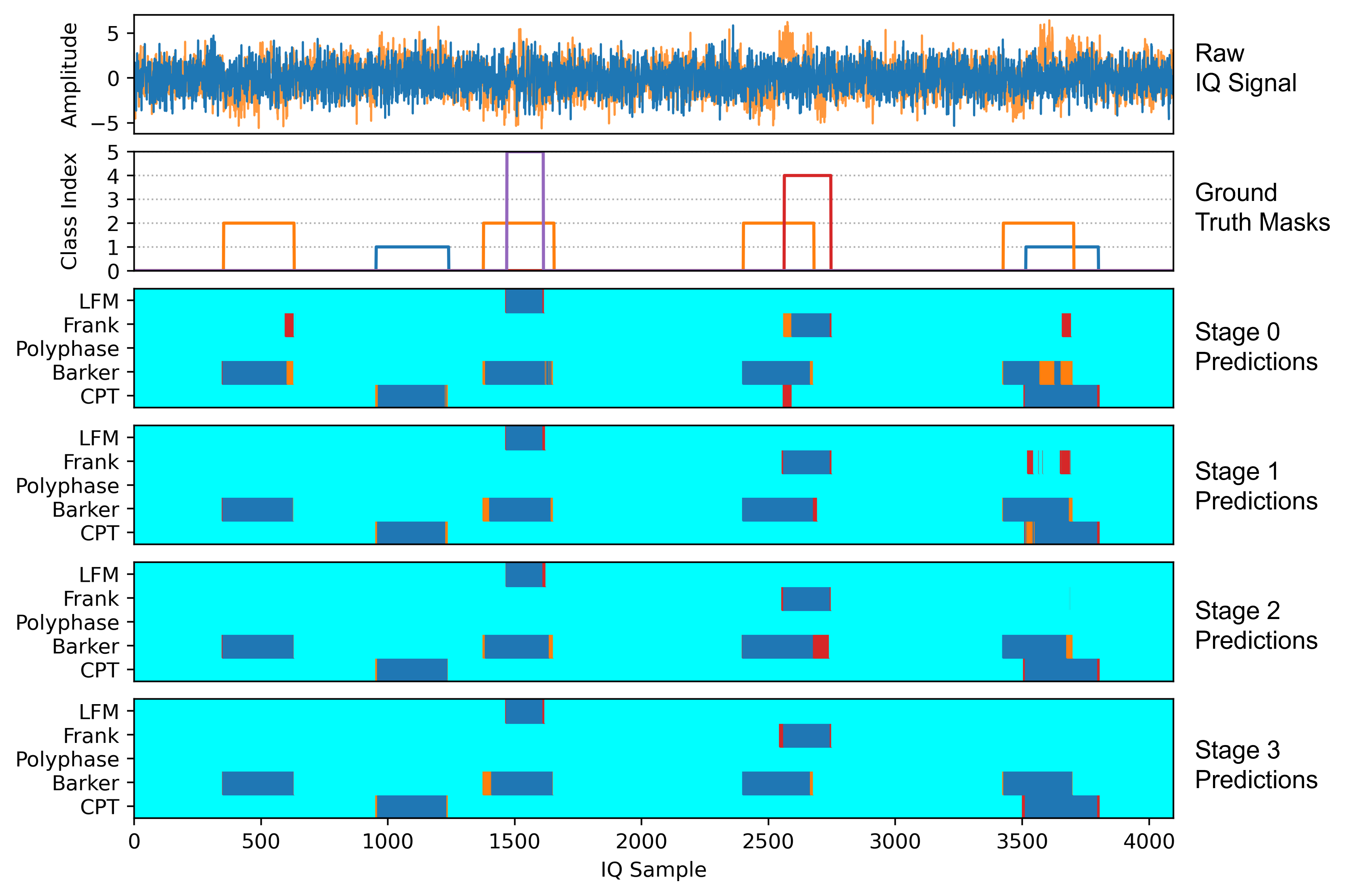}
    \caption{Segmentation predictions of the MS-UNet1D with $3$ stages at $-8$ \si{\deci\bel} SNR. The bar colours blue, cyan, red, and orange denote true positive, true negative, false positive, and false negative predictions, respectively.}
\label{fig:mask_preds}
\end{figure}

\begin{table*}[tb]
\caption{Comparison of segmentation models evaluated on the RadSeg test set. Each performance metric shows the average performance computed at $-20$, $-15$, $-10$, and $-5$ \si{\deci\bel} SNR. A higher value indicates better test performance.}
\label{table:results}
\vspace{8pt}
\def\arraystretch{0.8} 
\noindent\begin{tabular*}{\textwidth}
{@{\hspace{0.2cm}}@{\extracolsep{\stretch{1}}}*{14}{c|c|cccc|cccc|cccc}@{\hspace{0.2cm}}}
    \toprule
    \multicolumn{1}{l|}{\textbf{Model}} & \multicolumn{1}{c|}{\textbf{Stages}} & \multicolumn{4}{c|}{\textbf{F1@\{-20, -15, -10, -5\}}} & \multicolumn{4}{c|}{\textbf{Dice@\{-20, -15, -10, -5\}}} & \multicolumn{4}{c}{\textbf{IoU@\{-20, -15, -10, -5\}}} \\
    \midrule
    \multicolumn{1}{l|}{UNet1D} & - & 58.3 & 84.9 & 96.8 & 98.4 & 67.6 & 86.9 & 96.4 & 97.7 & 66.0 & 85.2 & 95.7 & 97.3 \\
    
    \multicolumn{1}{l|}{MS-UNet1D} & 1 & 65.8 & 89.2 & \textbf{98.1} & 99.1 & 77.4 & 93.1 & \textbf{98.9} & 99.2 & 76.3 & 92.0 & \textbf{98.5} & 99.0 \\
    
    \multicolumn{1}{l|}{MS-UNet1D} & 2 & \textbf{69.6} & \textbf{89.3} & 98.0 & \textbf{99.2} & \textbf{79.3} & \textbf{93.6} & 98.8 & \textbf{99.4} & \textbf{78.2} & \textbf{92.5} & 98.3 & \textbf{99.2} \\
    
    \midrule
    \multicolumn{1}{l|}{TCN} & - & 62.7 & 90.0 & 98.1 & 99.1 & \textbf{74.8} & 93.3 & 98.7 & 98.9 & \textbf{73.4} & 92.2 & 98.3 & 98.7 \\
    
    \multicolumn{1}{l|}{MS-TCN} & 1 & 64.4 & 91.1 & 98.0 & \textbf{99.4} & 73.2 & 93.8 & \textbf{99.0} & \textbf{99.6} & 71.9 & 92.7 & \textbf{98.6} & \textbf{99.4} \\
    
    \multicolumn{1}{l|}{MS-TCN} & 2 & \textbf{66.1} & \textbf{91.8} & \textbf{98.5} & 99.3 & 74.4 & \textbf{94.8} & 98.9 & 99.5 & 73.3 & \textbf{93.7} & 98.5 & 99.3 \\
    
    \midrule
    \multicolumn{1}{l|}{TCN++} & - & 67.5 & 91.5 & \textbf{98.3} & \textbf{99.0} & 78.8 & \textbf{95.1} & \textbf{98.7} & \textbf{99.0} & 77.9 & \textbf{94.1} & \textbf{98.3} & \textbf{98.8} \\
    
    \multicolumn{1}{l|}{MS-TCN++} & 1 & 71.7 & 90.9 & 97.7 & 98.8 & 79.3 & 93.7 & 97.9 & 98.8 & 78.3 & 92.6 & 97.4 & 98.6 \\
    
    \multicolumn{1}{l|}{MS-TCN++} & 2 & \textbf{74.4} & \textbf{91.6} & 97.7 & 98.8 & \textbf{79.7} & 94.7 & 98.4 & 98.8 & \textbf{78.8} & 93.8 & 97.9 & 98.5 \\
    \bottomrule
\end{tabular*}
\end{table*}

\begin{figure*}[tb]
\begin{minipage}[b]{0.245\linewidth}
  \centering
  \centerline{\includegraphics[width=1\linewidth]{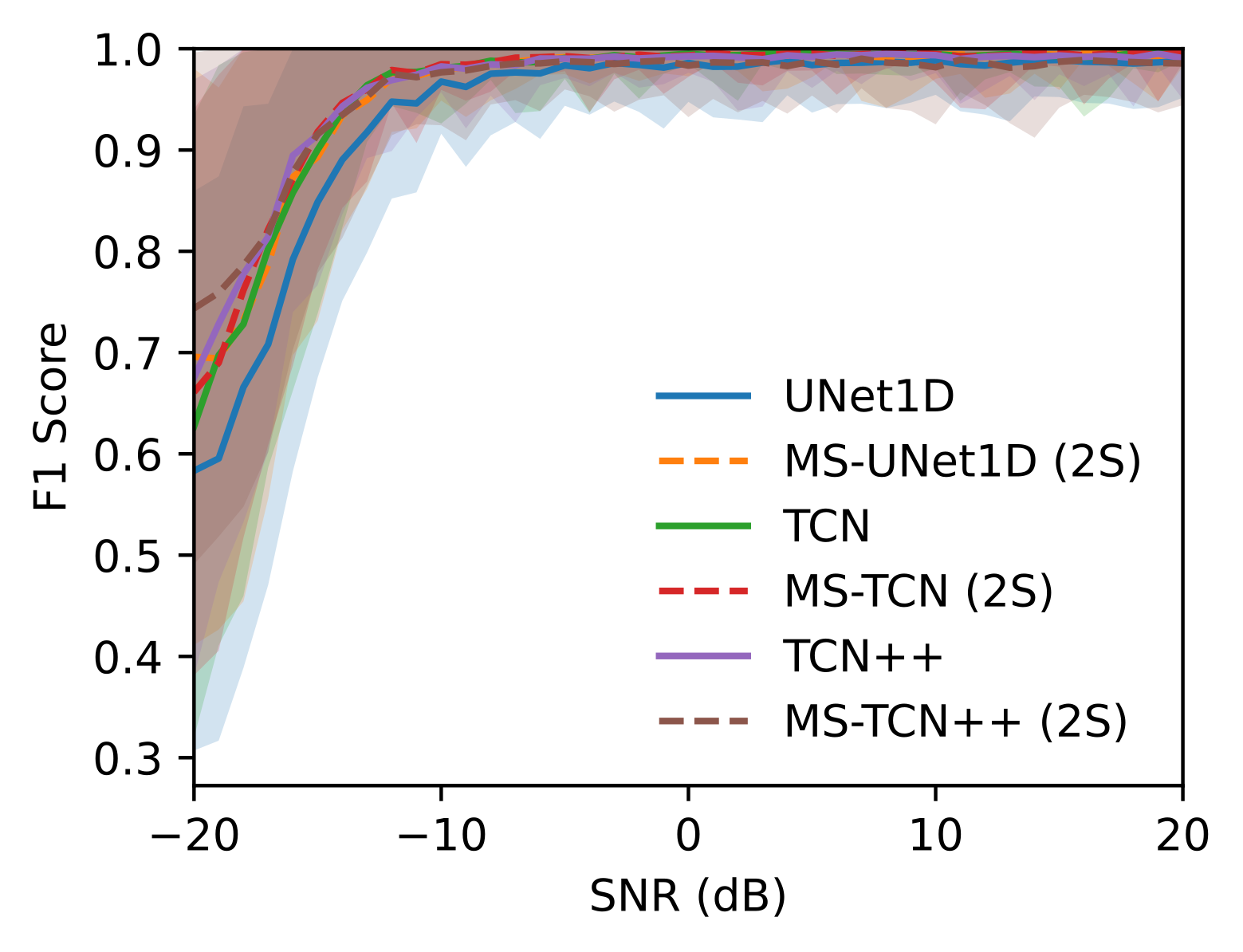}}
  \centerline{(a) Mean F1 score}\medskip
\end{minipage}
\hfill
\begin{minipage}[b]{0.245\linewidth}
  \centering
  \centerline{\includegraphics[width=1\linewidth]{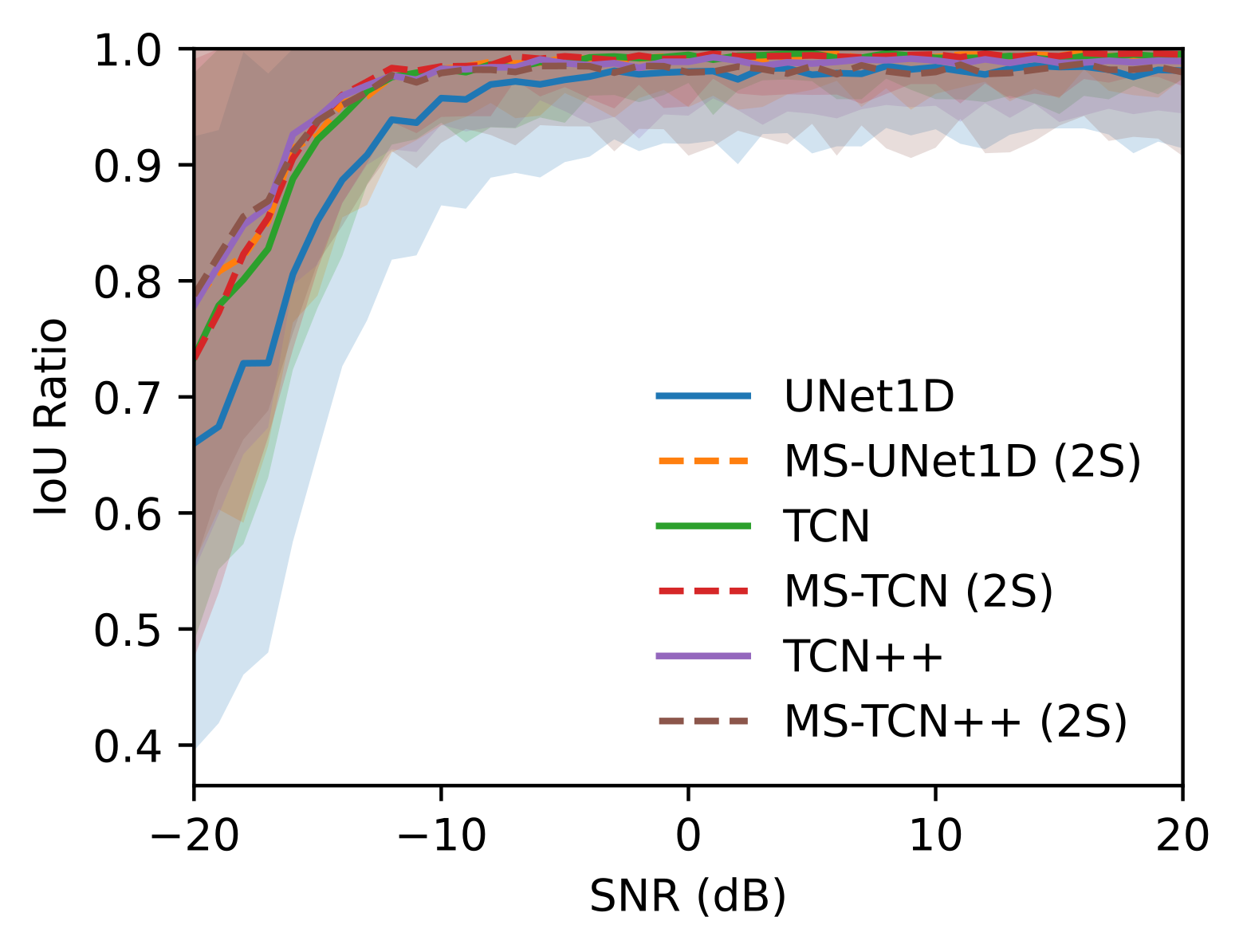}}
  \centerline{(b) Mean IoU ratio}\medskip
\end{minipage}
\hfill
\begin{minipage}[b]{0.245\linewidth}
  \centering
  \centerline{\includegraphics[width=1\linewidth]{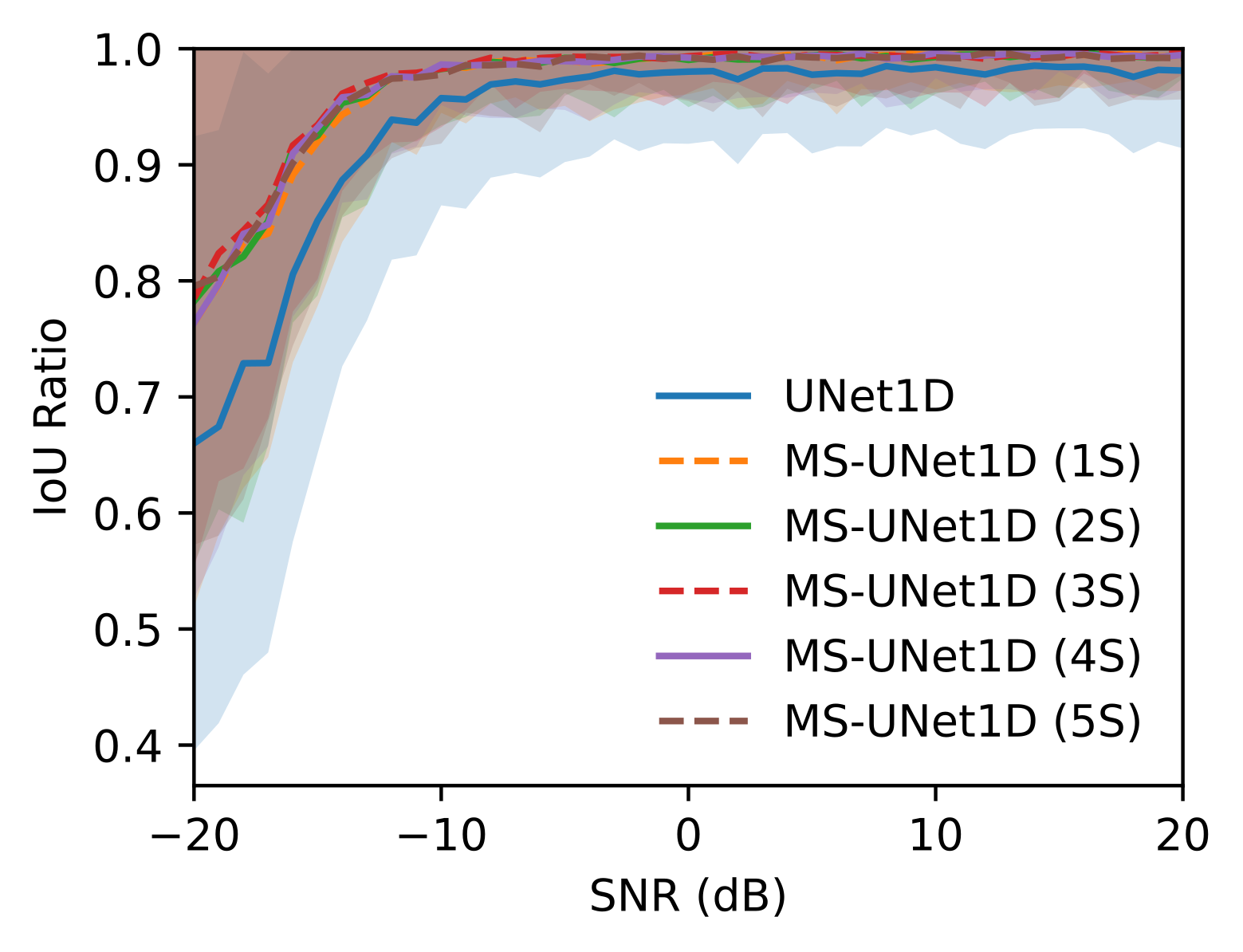}}
  \centerline{(c) MS-UNet1D}\medskip
\end{minipage}
\hfill
\begin{minipage}[b]{0.245\linewidth}
  \centering
  \centerline{\includegraphics[width=1\linewidth]{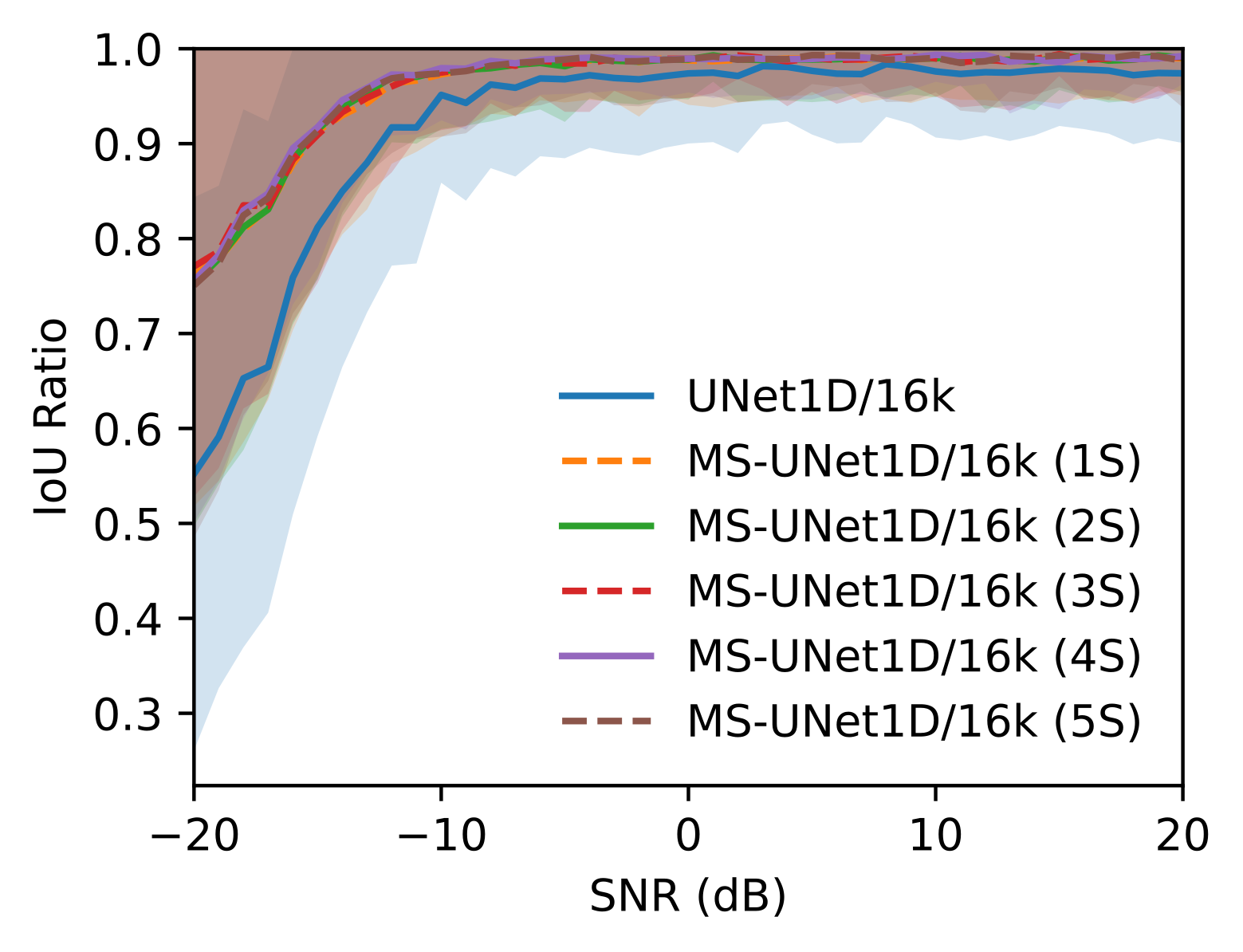}}
  \centerline{(d) MS-UNet1D/16k}\medskip
\end{minipage}
\hfill
\vspace{-5pt}
\caption{Test performance of multi-stage models across an SNR range of $-20$ to $20$ \si{\deci\bel}. Mean values are shown for each evaluation metric where the shaded regions represent the standard deviation of the respective metric.}
\label{fig:snr_plots}
\vspace{-5pt}
\end{figure*}

\vspace{-4pt}
\section{Experiments}
\label{sec:results}
\vspace{-5pt}

\subsection{Training Details}
\label{ssec:exp_details}
\vspace{-5pt}
We train and evaluate the models on a single Nvidia Tesla A100 GPU. Models are trained for $50$ epochs with a constant learning rate of $10^{-4}$ using the Adam optimiser. We standardise the raw IQ samples using the training population mean and variance. To improve generalisation, we apply data augmentation by sampling two random sets of $2 \times 4,096$ sequences from each IQ signal as inputs to the multi-stage model. While our models can process much longer sequences, this  was done to reduce the memory footprint in order to train efficiently on a single GPU. Using our configuration, training a UNet1D takes $3$ hours, while training TCN++ takes $15$ hours. Note that UNet1D and TCN++ contain approximately $10.8\text{m}$ and $23.1\text{m}$ model parameters, respectively.

\vspace{-5pt}
\subsection{Multi-stage Model Performance}
\label{ssec:model_perf}
\vspace{-5pt}
We evaluate the multi-stage models on RadSeg to establish a benchmark for radar pulse activity segmentation. For the test metrics, we consider the F1 score to assess sample-wise classification accuracy, while a channel-wise Dice coefficient and intersection-over-union (IoU) ratio are used to evaluate segmentation performance. A simple threshold of $0.5$ is used to binarise mask predictions to compute both the Dice coefficient and IoU ratio. The mean of each metric is computed for all predictions at each SNR. Note that correct predictions corresponding to 100\% true negative samples are neglected when computing the F1 score to prevent division by zero. 

Table \ref{table:results} provides a summary of results at various SNRs. Overall, all models perform exceptionally well for all metrics above -10 \si{\deci\bel}, while performance is poor at low SNRs. This is an expected trend which is consistent with similar radio signal recognition tasks \cite{jagannath_multi-task_2021-1,huang2023multi}. Without a multi-stage approach, the baseline UNet1D is outperformed by both TCN and TCN++ across all SNRs. A notable increase in segmentation performance can be observed across all models by including multiple stages. This performance gain is most significant for the MS-UNet1D at $-20$ \si{\deci\bel}, where a $15.5$\% increase in the IoU ratio is observed. This substantial improvement over the baseline UNet1D model is highlighted in Figure \ref{fig:snr_plots}, whereby the segmentation performance of the MS-UNet1D with only $2$ stages is on par with both TCN and MS-TCN++ at $-10$ \si{\deci\bel}. The effects of the multiple stages can be observed in the qualitative results show in Figure \ref{fig:mask_preds}. Each stage results in an incremental refinement of the channel-wise mask predictions. This underscores the benefits of multi-stage models for pulse activity segmentation whereby fine-grained signal features are preserved and incrementally refined, allowing the network to learn the higher-order positional relationships required to deinterleave and localise complex signal activities.

\vspace{-5pt}
\subsection{Ablation Study}
\label{ssec:ablation}
\vspace{-5pt}
We study the influence of various design considerations on the segmentation performance of MS-UNet1D. As indicated in Section \ref{ssec:model_perf}, increasing the number of stages significantly enhances segmentation performance across all SNRs, however there are diminishing returns as shown in Figure \ref{fig:snr_plots}(c). MS-UNet1D does not experience a notable slowdown during testing as the number of stages increases from $1$ to $5$. The inference speed of the model averages approximately $1.3$ $\si{\milli\second}$ in our experiments. While increasing the number of stages can be beneficial, it may lead to over-segmentation errors at low SNRs in locations where multiple signals co-exist. This is attributed to an imbalance in the occurrence of densely interleaved radar pulses, which are themselves rare in practice. We also experiment with the length of the feature vectors to observe its impact on the IoU ratio in Figure \ref{fig:snr_plots}(d). Increasing the length from $4,096$ to $16,384$ samples resulted in a slight drop in performance for MS-UNet1D across all SNRs. Lastly, we experiment with different loss functions for MS-UNet1D including BCE, Huber, and Dice loss, but did not find significant improvements across the test metrics.


\vspace{-5pt}
\section{Conclusion}
\label{sec:conclusion}
\vspace{-5pt}
This paper has presented a simple, yet highly effective multi-stage segmentation model for predicting fine-grained radar pulse activities in significantly degraded SNR environments. We created an open-source dataset containing $80,000$ long IQ sequences with complex interleaving radar signal characteristics, and provide precise multi-channel segmentation masks for each radar signal type. Our results demonstrate that through a multi-stage design, MS-UNet1D effectively retains fine-grained features and incrementally reduces segmentation errors. As a result, it achieves a substantial $15.5$\% increase in test performance (IoU) at $-20$ \si{\deci\bel} SNR and performs on par with MS-TCN++ while needing significantly fewer model parameters. In future work, the dataset may be extended to incorporate additional radar classes and behaviours to further investigate the practical utility of the proposed models.


\vspace{-5pt}
\section{Acknowledgement}
\label{sec:acknowledgement}
\vspace{-5pt}
The research for this paper received funding support from the Queensland Government through Trusted Autonomous Systems (TAS), a Defence Cooperative Research Centre funded through the Commonwealth Next Generation Technologies Fund and the Queensland Government.



\bibliographystyle{IEEEbib}
\bibliography{refs_icassp24}

\end{document}